\documentclass[runningheads]{llncs}
\usepackage{graphicx}
\usepackage{amsmath}
\usepackage{amssymb}
\usepackage{multirow}

\begin{document}
\title{Transformer-based Multi-Aspect Modeling for Multi-Aspect Multi-Sentiment Analysis}
\titlerunning{Transformer-based Multi-Aspect Modeling for MAMS}
%

\newcommand*\samethanks[1][\value{footnote}]{\footnotemark[#1]}
\author{Zhen Wu\thanks{Authors contributed equally.} \and
Chengcan Ying\samethanks \and
Xinyu Dai\thanks{Corresponding author.} \orcidID{0000-0002-4139-7337} \and
\\Shujian Huang \and
Jiajun Chen}
\authorrunning{Z. Wu et al.}
\institute{National Key Laboratory for Novel Software Technology, Nanjing University, Nanjing, 210023, China\\
\email{\{wuz,yingcc\}@smail.nju.edu.cn},
\email{\{daixinyu,huangsj,chenjj\}@nju.edu.cn}
}
\maketitle              

\begin{abstract}
Aspect-based sentiment analysis (ABSA) aims at analyzing the sentiment of a given aspect in a sentence. Recently, neural network-based methods have achieved promising results in existing ABSA datasets. However, these datasets tend to degenerate to sentence-level sentiment analysis because most sentences contain only one aspect or multiple aspects with the same sentiment polarity. To facilitate the research of ABSA, NLPCC 2020 Shared Task 2 releases a new large-scale Multi-Aspect Multi-Sentiment (MAMS) dataset. In the MAMS dataset, each sentence contains at least two different aspects with different sentiment polarities, which makes ABSA more complex and challenging. To address the challenging dataset, we re-formalize ABSA as a problem of multi-aspect sentiment analysis, and propose a novel Transformer-based Multi-aspect Modeling scheme (TMM), which can capture potential relations between multiple aspects and simultaneously detect the sentiment of all aspects in a sentence. Experiment results on the MAMS dataset show that our method achieves noticeable improvements compared with strong baselines such as BERT and RoBERTa, and finally ranks the 2nd in NLPCC 2020 Shared Task 2 Evaluation.

\end{abstract}

\section{Introduction}
\begin{figure}[!htbp]
	\centering
	\includegraphics[width=0.7\textwidth]{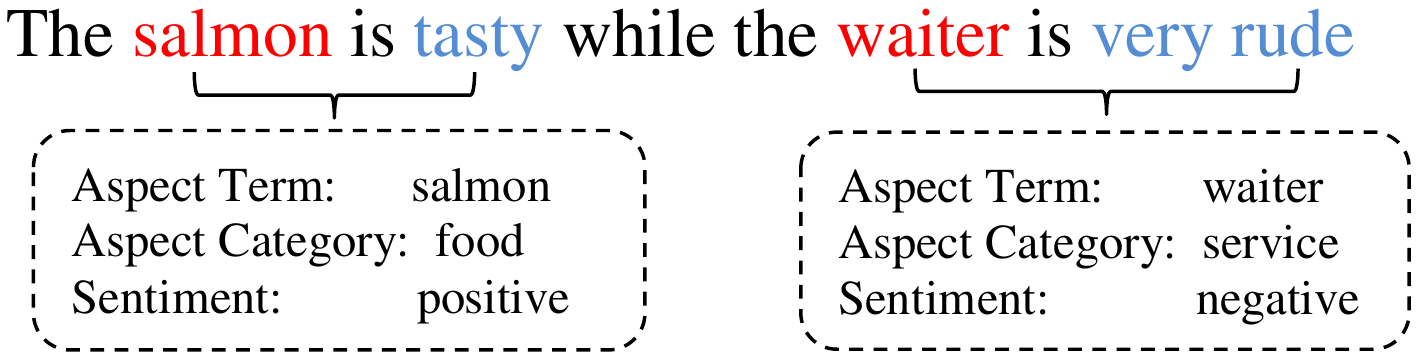}
	\caption{An example of the ATSA and ACSA subtasks. The terms in red are two given aspect terms. Note that the annotations for ATSA and ACSA can be separated.}
	\label{sentenceexmaple}
\end{figure}

Aspect-based sentiment analysis (ABSA) is a fine-grained sentiment analysis task, which aims to detect the sentiment polarity towards one given aspect in a sentence~\cite{DBLP:series/synthesis/2012Liu,DBLP:journals/ftir/PangL07,DBLP:conf/semeval/PontikiGPPAM14}. The given aspect usually refers to the aspect term or the aspect category. An aspect term is a word or phrase explicitly mentioned in the sentence representing the feature or entity of products or services. Aspect categories are pre-defined coarse-grained aspect descriptions, such as \emph{food}, \emph{service}, and \emph{staff} in restaurant review domain. Therefore, ABSA contains two subtasks, namely Aspect Term Sentiment Analysis (ATSA) and Aspect Category Sentiment Analysis (ACSA). Fig.~\ref{sentenceexmaple} shows an example for ATSA and ACSA. Given the sentence ``\emph{The salmon is tasty while the waiter is very rude}'',  the sentiments toward the two aspect terms ``\emph{salmon}'' and ``\emph{waiter}'' are respectively positive and negative. ACSA is to detect the sentiment polarity towards the given pre-defined aspect category, which is explicitly or implicitly expressed in the sentence. There are two aspect categories in the sentence of Fig.~\ref{sentenceexmaple}, i.e., \emph{food} and \emph{waiter}, and their sentiments are respectively positive and negative. Note that the annotations for ATSA and ACSA can be separated.

To study ABSA, several public datasets are constructed, including multiple SemEval Challenges datasets~\cite{DBLP:conf/semeval/PontikiGPAMAAZQ16,DBLP:conf/semeval/PontikiGPMA15,DBLP:conf/semeval/PontikiGPPAM14} and Twitter dataset~\cite{DBLP:conf/acl/DongWTTZX14}. However, in these datasets, most sentences consist of only one aspect or multiple aspects with the same sentiment polarity, which makes ABSA degenerate to sentence-level sentiment analysis~\cite{DBLP:conf/emnlp/JiangCXAY19}. For example, there are only 0.09\% instances in Twitter dataset belonging to the case of multi-aspects with different sentiment polarities. To promote the research of ABSA, NLPCC 2020 Shared Task 2 releases a Multi-Aspect Multi-Sentiment (MAMS) dataset. In the MAMS dataset, each sentence consists of at least two aspects with different sentiment polarities. Obviously, the property of multi-aspect multi-sentiment makes the proposed dataset more challenging compared with existing ABSA datasets.

To deal with ABSA, recent works employ neural networks and achieve promising results in previous datasets, such as attention networks~\cite{DBLP:conf/emnlp/WangHZZ16,DBLP:conf/ijcai/MaLZW17,DBLP:conf/emnlp/FanFZ18}, memory networks~\cite{DBLP:conf/emnlp/TangQL16,DBLP:conf/emnlp/ChenSBY17}, and BERT~\cite{DBLP:conf/emnlp/JiangCXAY19}. These works separate multiple aspects of a sentence into several instances and process one aspect each time. As a result, they only consider local sentiment information for the given aspect while neglecting the sentiments of other aspects in the same sentence as well as the relations between multiple aspects. This setting is unsuitable, especially for the new MAMS dataset, as multiple aspects of a sentence usually have different sentiment polarities in the MAMS dataset, and knowing sentiment of a certain aspect can help infer sentiments of other aspects. To address the issue, we re-formalize ABSA as a task of multi-aspect sentiment analysis, and propose a \textbf{T}ransformer-based \textbf{M}ulti-aspect \textbf{M}odeling method (TMM) to simultaneously detect the sentiment polarities of all aspects in a sentence. Specifically, we adopt the pre-trained RoBERTa~\cite{DBLP:journals/corr/abs-1907-11692} as backbone network and build a multi-aspect scheme for MAMS based on transformer~\cite{DBLP:conf/nips/VaswaniSPUJGKP17} architecture, then employ multi-head attention to learn the sentiment and relations of multi-aspects. Compared with existing works, our method has three advantages:

\begin{enumerate}
	\item It can capture sentiments of all aspects synchronously in a sentence and relations between them, thereby avoid focusing on sentiment information belonging to other aspects mistakenly.
	
	\item Modeling multi-aspect simultaneously can improve computation efficiency largely without additional running resources.
	
	\item Our method applies the strategy of transfer learning, which exploits large-scale pre-trained semantic and syntactic knowledge to benefit the downstream MAMS task.
	
\end{enumerate}

Finally, our proposed method obtains obvious improvements for both ATSA and ACSA  in the MAMS dataset, and rank the second place in the NLPCC 2020 Shared Task 2 Evaluation.

\section{Proposed Method}
In this section, we first re-formalize the ABSA task, then present our proposed Transformer-based Multi-aspect Modeling scheme for ATSA and ACSA. The final part introduces the fine-tuning and training objective.

\subsection{Task Formalization}

Prior studies separate multiple aspects and formalize ABSA as a problem of sentiment classification toward one given aspect $a$ in the sentence $s=\{w_1, w_2, \cdots, w_n\}$. In ATSA, the aspect term $a$ is a span of the sentence $s$ representing the feature or entity of products or services. For ACSA, the aspect category $a\in A$ and $A$ is the pre-defined aspect set, i.e., \{\emph{food}, \emph{service}, \emph{staff}, \emph{price}, \emph{ambience}, \emph{menu}, \emph{place}, \emph{miscellaneous}\} for the new MAMS dataset. The goal of ABSA is to assign a sentiment label $y\in C$ to the aspect $a$ of the sentence $s$, where $C$ is the set of sentiment polarities (i.e., postive, neural and negative).

In this work, we re-formalize ABSA as a task of multi-aspect sentiment classifcation. Given a sentence $s=\{w_1, w_2, \cdots, w_n\}$ and $m$ aspects $\{a_1, a_2, \cdots, a_m\}$ mentioned in $s$, the objective of MAMS is to simultaneously detect the sentiment polarities $\{y_1, y_2, \cdots, y_m\}$ of all aspects $\{a_1, a_2, \cdots, a_m\}$, where $y_i$ corresponds to the sentiment label of the aspect $a_i$.

\subsection{Transformer-based Multi-Aspect Modeling for ATSA}
\begin{figure}[!htbp]
	\centering
	\includegraphics[width=0.9\textwidth]{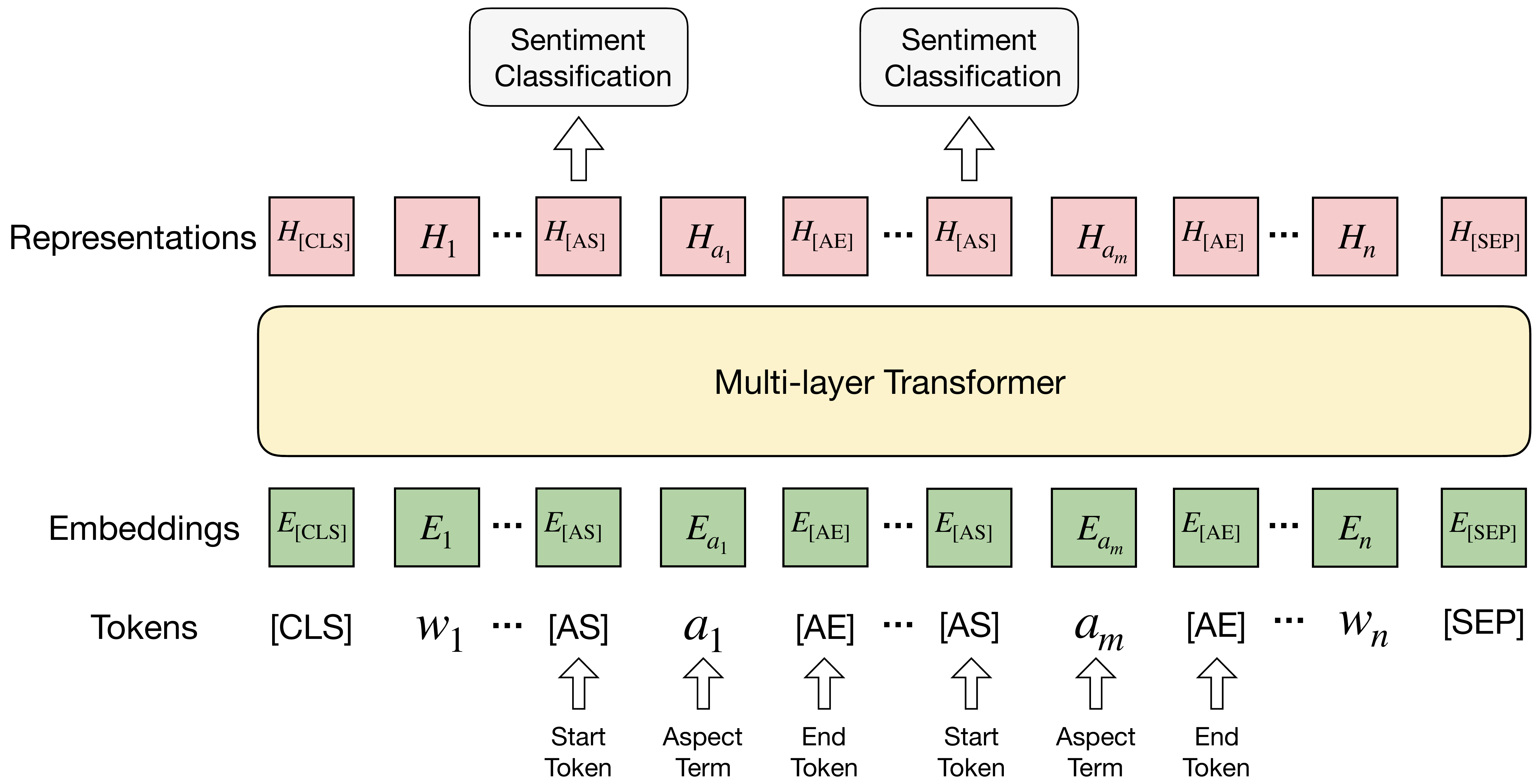}
	\caption{Transformer-based Multi-Aspect Modeling for ATSA. In the above example, the aspect $a_i$ may contain multiple words, and each word of the sentence might be split into several subwords. For simplicity, here we do not represent them with subword tokens.}
	\label{atsamodelfig}
\end{figure}

Recently, Bidirectional Encoder Representations from Transformers (BERT)~\cite{DBLP:conf/naacl/DevlinCLT19} achieves great success by pre-training a language representation model on large-scale corpora then fine-tuning on downstream tasks. When fine-tuning on classification tasks, BERT uses the specific token {\tt[CLS]} to obtain task-specific representation, then applies one additional output layer for classification. For ABSA, previous work concatenates the given single aspect and the original sentence as the input of BERT encoder, then leverages the representation of {\tt[CLS]} for sentiment classification~\cite{DBLP:conf/emnlp/JiangCXAY19}.

Inspired by BERT, we design a novel Transformer-based Multi-Aspect Modeling scheme (TMM) to address MAMS task with simultaneously detecting the sentiments of all aspects in a sentence. Here we take ATSA subtask as example to elaborate on it. Specifically, given a sentence $\{w_1, \cdots, a_1, \cdots, a_m, \cdots, w_n\}$, where the aspect terms are denoted in the original sentence for the ease of following description, we propose two specific tokens {\tt[AS]} and {\tt[AE]} to respectively represent the start position and end position of aspect in the sentence. With the two tokens, the original sentence $\{w_1, \cdots, a_1, \cdots, a_m, \cdots, w_n\}$ can be transformed into the sequence $\{w_1, \cdots,  \text{{\tt[AS]}}, a_1, \text{{\tt[AE]}}, \cdots, \text{{\tt[AS]}}, a_m, \text{{\tt[AE]}}, \cdots, w_n\}$. Based on this new input sequence, we then employ multi-layer transformer to automatically learn the sentiments and relations between multiple aspects.

As shown in Fig.~\ref{atsamodelfig}, we finally fetch the representation $\mathbf{H}_{\rm{[AS]}}$ of the start token {\tt[AS]} of each aspect as feature vector to classify the sentiment of aspect.

\subsection{Transformer-based Multi-Aspect Modeling for ACSA}
\begin{figure}[!htbp]
	\centering
	\includegraphics[width=0.9\textwidth]{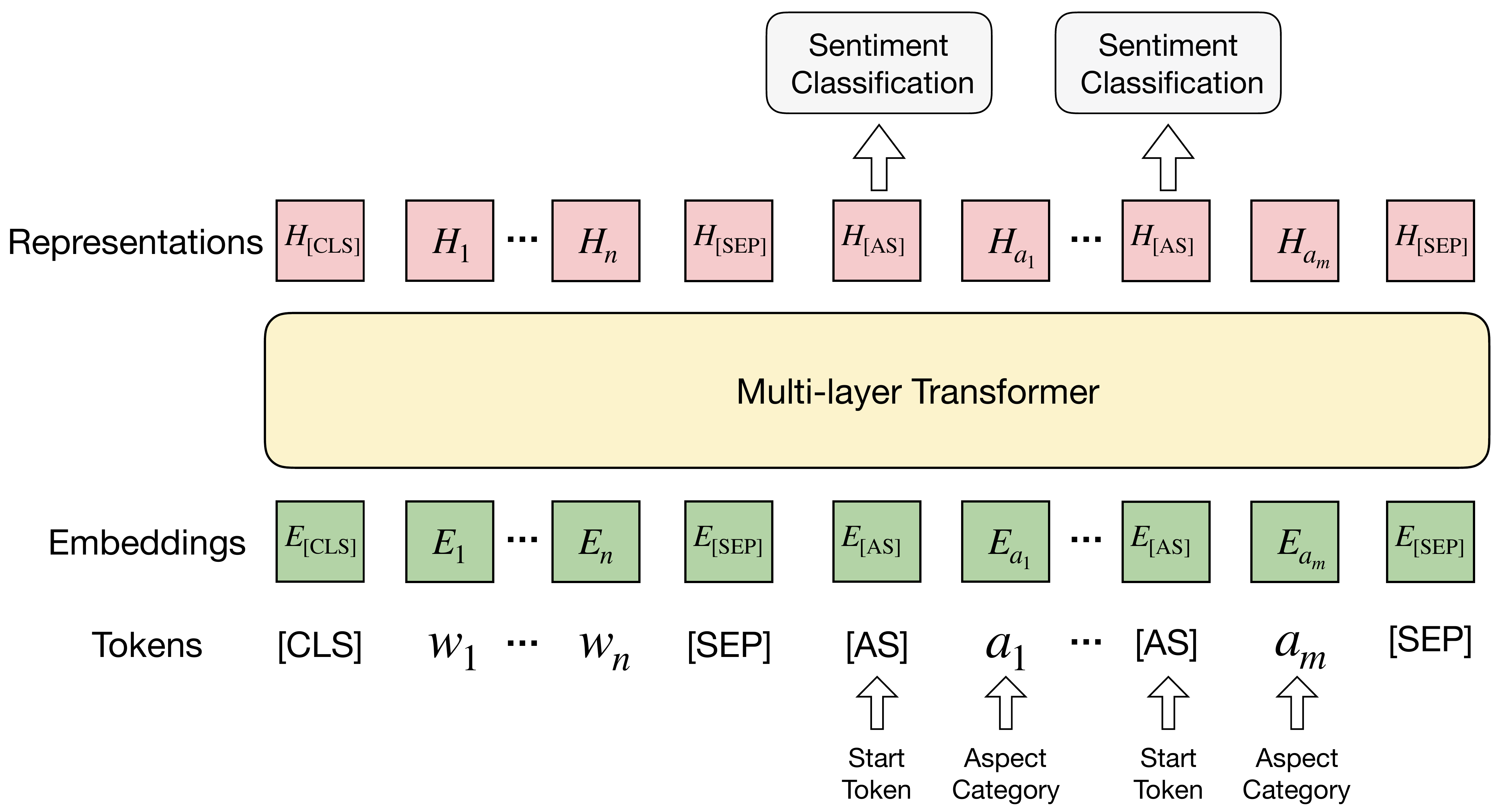}
	\caption{Transformer-based Multi-Aspect Modeling for ACSA.}
	\label{acsamodelfig}
\end{figure}

Since aspect categories are pre-defined and may be not mentioned explicitly in the sentence, the above TMM scheme needs some modifications for ACSA. Given the sentence $s=\{w_1, w_2, \cdots, w_n\}$ and aspect categories $\{a_1, a_2, \cdots, a_m\}$ in $s$, we concatenate the sentence and aspect categories, and only use the token {\tt [AS]} to separate multiple aspects because each aspect category is a single word, finally forming the input sequence $\{w_1, w_2, \cdots, w_n, \text{\tt [AS]},  a_1, \text{\tt [AS]}, a_2, \cdots, \text{\tt [AS]}, a_m\}$. As Fig.~\ref{acsamodelfig} shows, after multi-layer transformer, we use the representation $\mathbf{H}_{\rm{[AS]}}$ the indication token {\tt [AS]} of each aspect category for sentiment classifcation.

\subsection{Fine-tuning and Training Objective}

As aforementioned, we adopt the pre-trained RoBERTa as backbone network, then fine-tune it on the MAMS dataset with the proposed TMM scheme. RoBERTa is a robustly optimized BERT approach and pre-trained with the larger corpora and batch size.

When in the fine-tuning stage, we employ a softmax classifier to map the representation $\mathbf{H}^i_{\rm{[AS]}}$ of aspect $a_i$ into the sentiment distribution $\hat {\mathbf{y}}_i$ as follow:
\begin{equation}
\hat {\mathbf{y}}_i=\mathrm{softmax}(\mathbf{W}_o\mathbf{H}^i_{\rm{[AS]}}+\mathbf{b}_o),
\end{equation}
where $\mathbf{W}_o$ and $\mathbf{b}_o$ respectively denote weight matrix and bias.

Finally, we use cross-entropy loss between predicted sentiment label and the golden sentiment label as training loss, which is defined as follows:
\begin{equation}
Loss=-  \sum_{s\in D}\sum_{i=1}^{m}\sum_{j\in C}\mathbb{I}(y_i=j) \log \hat y_{i,j},
\end{equation}
where $s$ and $D$ respectively denote a sentence and training dataset, $m$ represents the number of aspects in the sentence $s$, $C$ is the sentiment label set, $y_i$ denotes the ground truth sentiment of aspect $a_i$ in $s$, and $\hat y_{i,j}$ is the predicted probability of the $j$-th sentiment towards the aspect $a_i$ in the input sentence.

\section{Experiment}

\subsection{Dataset and Metrics}
\begin{table}[!htbp]
		\caption{Statistics of the MAMS dataset. Sen. and Asp. respectively denotes the numbers of sentences and given aspects in the dataset. Ave. represents the average number of aspects in each sentence. Pos., Neu. and Neg. respectively indicate the numbers of positive, neutral and negative sentiment.}
		\label{datasets}
		\centering
		\begin{tabular}{cc|ccc|ccc}
			\hline
			\multicolumn{2}{c|}{Datasets} & {Sen.} & {Asp.} & {Ave.} & {Pos.} & {Neu.} & {Neg.}\\
			\hline
			\multirow{3}*{ATSA}  & Train & 4297 & 11186 & 2.60 & 3380 & 5042 & 2764 \\
			& Dev & 1000 & 2668 & 2.67 & 803 & 1211 & 654\\
			& Test & 1025 & 2676 & 2.61 & 1046 & 1085 & 545\\
			\hline
			\multirow{3}*{ACSA}  & Train & 3149 & 7090 & 2.25 & 1929 & 3077 & 2084\\
			& Dev & 800 & 1789 & 2.24 & 486 & 781 & 522\\
			& Test & 684 &  1522 & 2.23 & 562 & 612 & 348\\
			\hline
		\end{tabular}
\end{table}

Similar to SemEval 2014 Restaurant Review dataset~\cite{DBLP:conf/semeval/PontikiGPPAM14}, the original sentences in NLPCC 2020 Shared Task 2 are from  the Citysearch New York dataset~\cite{DBLP:conf/webdb/GanuEM09}. Each sentence is annotated with three experienced researchers working on natural language processing. In the released MAMS dataset, the annotations for ATSA and ACSA are separated. For ACSA, they pre-defined eight coarse-grained aspect categories, i.e., \emph{food}, \emph{service}, \emph{staff}, \emph{price}, \emph{ambience}, \emph{menu}, \emph{place}, and \emph{miscellaneous}. The sentences consisting of only one aspect or multiple aspects with the same sentiment polarities are deleted, thus each sentence at least contains two aspects with different sentiments. This property makes the MAMS dataset more challenging. The statistics of the MAMS dataset are shown in Tabel~\ref{datasets}.

NLPCC 2020 Shared Task 2 uses Macro-F1 to evaluate the performance of different systems, which is calculated as follows:
\begin{align}
Precision (P) &=  TP / (TP+FP),\\
Recall (R) &= TP / (TP+FN),\\
F1 &= 2*P*R / (P+R),
\end{align}
where TP represents true positives, FP represents false positives, TN represents true negatives, and FN represents false negatives. Macro-F1 value is the average of F1 value of each category. The final evaluation result is the average result of Macro-F1 values on the two subtasks (i.e., ATSA and ACSA). In this work,  we also use standard Accuracy as the metric to evaluate different methods.

\subsection{Experiment Settings}
We use pre-trained RoBERTa as backbone network, then fine-tune it on downstream ATSA or ACSA subtask with our proposed Transformer-based Multi-aspect Modeling scheme. The RoBERTa has 24 layers of transformer blocks, and each block has 16 self-attention heads. The dimension of hidden size is 1024. When fine-tuning on ATSA or ACSA, we apply Adam optimizer~\cite{DBLP:journals/corr/KingmaB14} to update model parameters. The initial learning rate is set to 1e-5, and the mini-batch size is 32. We use the official validation set for hyperparameters tuning. Finally, we run each model 3 times and report the average results on the test set.

\subsection{Compared Methods}
To evaluate the performance of different methods, we compare our RoBERTa-TMM method with the following baselines on ATSA and ACSA.

\begin{itemize}
	\item \textbf{LSTM}: We use the vanilla LSTM to encode sentence and apply the average of all hidden states for sentiment classification.
	
	\item \textbf{TD-LSTM}: TD-LSTM~\cite{DBLP:conf/coling/TangQFL16} employs two LSTM networks respectively to encode the left context and right context of the aspect term, then concatenates them for sentiment classification.
	
	\item \textbf{AT-LSTM}: AT-LSTM~\cite{DBLP:conf/emnlp/WangHZZ16} uses the aspect representation as query, and employs the attention mechanism to capture aspect-specific sentiment information. For ATSA, the aspect term representation is the average of word vectors in the aspect term. For ACSA, the aspect category representation is randomly initialized and learned in the training stage.
	
	\item \textbf{ATAE-LSTM}: ATAE-LSTM~\cite{DBLP:conf/emnlp/WangHZZ16} is an extension of AT-LSTM. It concatenates the aspect representation and word embedding as the input of LSTM.
	
	\item \textbf{BiLSTM-Att}: BiLSTM-Att is our implemented model similar to AT-LSTM, which uses bidirectional LSTM to encode the sentence and applies aspect attention to capture the aspect-dependent sentiment.
	
	\item \textbf{IAN}: IAN~\cite{DBLP:conf/ijcai/MaLZW17} applies two LSTM to respectively encode the sentence and aspect term, then proposes the interactive attention to learn representations of the sentence and aspect term interactively. Finally, the two representations are concatenated for sentiment prediction. 
	
	\item \textbf{RAM}: RAM~\cite{DBLP:conf/emnlp/ChenSBY17} employs BiLSTM to build memory and then applies GRU-based multi-hops attention to generate the aspect-dependent sentence representation for predicting the sentiment of the given aspect.
	
	\item \textbf{MGAN}: MGAN~\cite{DBLP:conf/emnlp/FanFZ18}  proposes fine-grained attention mechanism to capture the word-level interaction between aspect term and context, then combines it with coarse-grained attention for ATSA.
	
\end{itemize}

In addition, we also compare strong transformer-based models including $\text{BERT}_\text{BASE}$ and RoBERTa. They adopt the conventional ABSA scheme and deal with one aspect each time.
\begin{itemize}
	\item $\textbf{BERT}_\textbf{BASE}$: $\text{BERT}_\text{BASE}$~\cite{DBLP:conf/naacl/DevlinCLT19} has 12 layers transformer blocks, and each block has 12 self-attention heads. When fine-tuning for ABSA, it concatenates the aspect and the sentence to form segment pair, then use the representation of the {\tt [CLS]} token after multi-layer transformers for sentiment classification.
	
	\item \textbf{RoBERTa}: RoBERTa~\cite{DBLP:journals/corr/abs-1907-11692} is a robustly optimized BERT approach. It replaces the static masking in BERT with dynamic masking, removes the next sentence prediction, and pre-trains with larger batches and corpora.
	
\end{itemize}

\subsection{Main Results and Analysis}
\begin{table}[!htbp]
	\centering
	\caption{Main experiment results on ATSA and ASCA (\%). The results with the marker $^*$ are from official evaluation and they do not provide accuracy performance.}
	\label{table_main_results}
	\begin{tabular}{c|cc|cc}
		\hline
		\multirow{2}{*}{Model} & \multicolumn{2}{c|}{ATSA}       & \multicolumn{2}{c}{ACSA}                        \\ \cline{2-5} 
		& Acc.           & F1             & Acc.              & F1.               \\ \hline
		LSTM              & 48.45         & 47.45          & 45.86              & 45.04            \\
		BiLSTM-Att              &71.91          & 71.04          & 69.84              & 69.28             \\
		TD-LSTM				&75.01          & 73.80          & -              & -            \\
		AT-LSTM              &67.95          & 66.87          & 68.39              & 67.98             \\
		ATAE-LSTM              & 65.26          & 64.48          & 66.41              & 66.15             \\
		IAN              & 70.02          & 68.88          & -              & -             \\
		RAM              & 75.58          & 74.46          & -              & -             \\
		MGAN             & 75.37          & 74.40          & -              & -             \\
		\hline
		$\text{BERT}_\text{BASE}$              & 82.12          & 81.29          & 72.88              &     72.91        \\
		RoBERTa              &   83.71       &     83.17    & 77.44              &      77.29       \\		\hline
		RoBERTa-TMM              & {85.64}          & {85.08}          &        {78.03}       &        {77.79}     \\
		$\text{RoBERTa-TMM}_{ensemble}$           &   -       &     85.24$^*$    & -            &      79.41$^*$       \\		
	 \hline
	\end{tabular}
\end{table}

Table~\ref{table_main_results} gives the results of different methods on two subtasks of ABSA.

The first part shows the performance of non-transformer-based baselines. We can observe that the vanilla LSTM performs very pool in this new MAMS dataset, because it does not consider any aspect information and is a sentence-level sentiment classification model. In fact, LSTM can obtain pretty good results on previous ABSA datasets, which reveals the challenge of the MAMS dataset. Compared with other attention-based models, RAM and MGAN achieve better performance on ATSA, which validates the effectiveness of multi-hops attention and multi-grained attention for detecting the sentiment of aspect. It is surprising that the TD-LSTM obtains competitive results among non-transformer-based baselines. This result indicates that modeling position information of aspect term may be crucial for the MAMS dataset.

The second part gives two strong baselines, i.e., $\text{BERT}_\text{BASE}$ and RoBERTa. They follow the conventional ABSA scheme and deal with one aspect each time. It is observed that they outperform the non-transformer-based models significantly, which shows the power of pre-trained language models. Benefiting from the larger datasets, batch size and the more parameters,  RoBERTa obtains better performance than $\text{BERT}_\text{BASE}$ on ATSA and ACSA.

Compared with the strongest baseline RoBERTa, our proposed Transformer-based Multi-aspect Modeling method RoBERTa-TMM still achieves obvious improvements in the challenging MAMS dataset. Specifically, it outperforms RoBERTa by 1.93\% and 1.91\% respectively in accuracy and F1-score for ATSA. In terms of ACSA, the improvement of RoBERTa-TMM against RoBERTa is relatively limited. This may be attributed to that the predefined aspect categories are abstract and it is challenging to find their corresponding sentiment spans from the sentence even in the multi-aspect scheme. Nevertheless, the improvement in ACSA is still substantial because the data size of the MAMS dataset is sufficient and even large-scale for ABSA research. Finally, our RoBERTa-TMM-based ensemble system achieves 85.24\% and 79.41\% respectively for ATSA and ACSA in F1-score, and ranks the 2nd in NLPCC 2020 Shared Task 2 Evaluation.

\subsection{Case Study}
\begin{figure}[!htbp]
	\centering
	\includegraphics[width=0.9\textwidth]{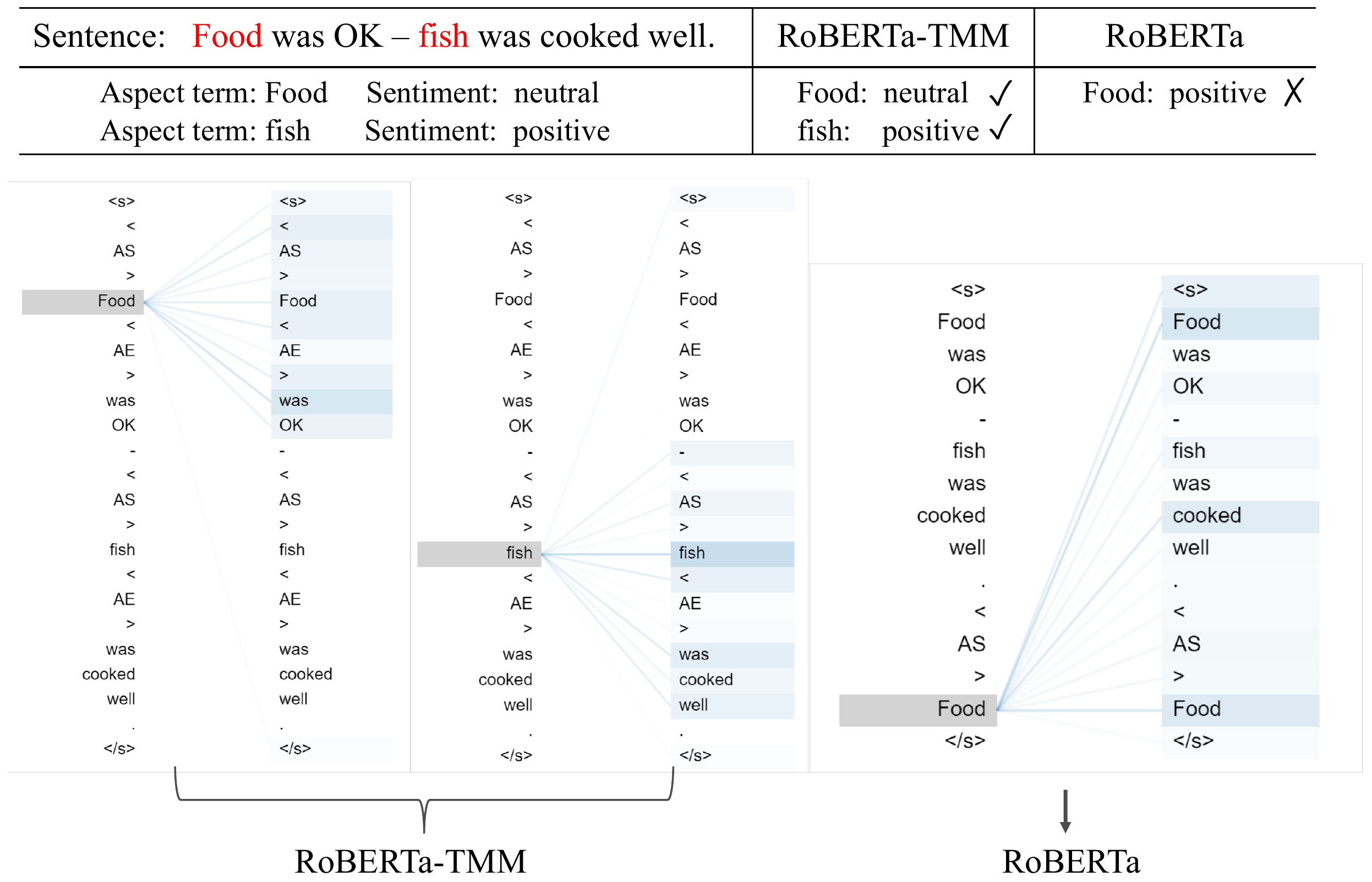}
	\caption{Attention visualization of RoBERTa-TMM and RoBERTa in ATSA. The words in red are two given aspect terms. The darker blue denotes the bigger attention weight.}
	\label{casestudy}
\end{figure}

To further validate the effectiveness of the proposed TMM scheme, we take a sentence from ATSA as example, and average the attention weight of different heads in RoBERTa-TMM and RoBERTa models, finally visualize them in Fig.~\ref{casestudy}.

From the results of attention visualization, we can see that the two aspect terms in the RoBERTa-TMM model capture the corresponding sentiment spans correctly through multi-aspect modeling. In contrast, given the aspect term ``\emph{Food}'', RoBERTa mistakenly focuses on the sentiment spans of the other aspect term ``\emph{fish}'' due to lacking other aspects information, thus making wrong sentiment prediction. The attention visualization indicates that the  RoBERTa-TMM can detect the corresponding sentiment spans of different aspects and avoid wrong attention as much as possible by simultaneously modeling multi-aspect and considering the potential relations between multiple aspects.

\section{Related Work}

\subsection{Aspec-based Sentiment Analysis}
Aspect-based sentiment analysis (ABSA) has been studied in the last decade. Early works devote to designing effective hand-crafted features, such as n-gram features~\cite{DBLP:conf/acl/JiangYZLZ11,DBLP:conf/semeval/KiritchenkoZCM14} and sentiment lexicons~\cite{DBLP:conf/ijcai/VoZ15}. Motivated by the success of deep learning in many tasks~\cite{DBLP:conf/nips/KrizhevskySH12,DBLP:journals/taslp/DahlYDA12,DBLP:conf/nips/BengioDV00}, recent works adopt neural network-based methods to automatically learn low-dimension and continuous features for ABSA.~\cite{DBLP:conf/coling/TangQFL16} separates the sentence into the left context and right context according to the aspect term, then employs two LSTM networks respectively to encode them from the two sides of sentence to the aspect term. To capture aspect-specific context,~\cite{DBLP:conf/emnlp/WangHZZ16} proposes the aspect attention mechanism to aggregate important sentiment information from the sentence toward the given aspect. Following the idea,~\cite{DBLP:conf/ijcai/MaLZW17} introduces the interactive attention networks (IAN) to learn attentions in context and aspect term interactively, and generates the representations for aspect and context words separately. Besides, some works employ memory network to detect more powerful sentiment information with multi-hops attention and achieve promising results~\cite{DBLP:conf/emnlp/TangQL16,DBLP:conf/emnlp/ChenSBY17}. Instead of the recurrent network,~\cite{DBLP:conf/acl/LiX18} proposes the aspect information as the gating mechanism based on convolutional neural network, and dynamically selects aspect-specific information for aspect sentiment detection. Subsequently, BERT based method achieves state-of-the-art performance for the ABSA task~\cite{DBLP:conf/emnlp/JiangCXAY19}.

However, the above methods perform ABSA with the conventional scheme that separates multiple aspects in the same sentence and analyzes one aspect each time. They only consider local sentiment information for the given aspect and possibly focus on sentiment information belonging to other aspects mistakenly. In contrast, our proposed Transformer-based Multi-aspect Modeling scheme (TMM) aims to learn sentiment information and relations between multiple aspects for better prediction. 

\subsection{Pre-trained Language Model}
Recently, substantial works have shown that pre-trained language models can learn universal language
representations, which are beneficial for downstream NLP tasks and can avoid training a new model from scratch~\cite{DBLP:conf/naacl/DevlinCLT19,DBLP:conf/nips/YangDYCSL19,DBLP:conf/iclr/LanCGGSS20,DBLP:journals/corr/abs-1907-11692}. These pre-trained models, e.g., GPT, BERT, XLNet, RoBERTa, use the strategy of first pre-training then fine-tuning and achieve the great success in many NLP tasks. To be specific, they first pre-train some self-supervised objectives, such as the masked language model (MLM), next sentence prediction (NSP), or sentence order prediction (SOP)~\cite{DBLP:conf/iclr/LanCGGSS20} on the large corpora, to learn complex semantic and syntactic pattern from raw text. When fine-tuning on downstream tasks, they generally employ one additional output layer to learn task-specific knowledge. 

Following the successful learning paradigm, in this work, we employ RoBERTa as the backbone network, then fine-tune it with the TMM scheme on the MAMS dataset to perform ATSA and ACSA.

\section{Conclusion}
Facing the challenging MAMS dataset, we re-formalize ABSA as a task of multi-aspect sentiment analysis in this work and propose a novel Transformer-based Multi-aspect Modeling scheme (TMM) for MAMS, which can determine the sentiments of all aspects in a sentence simultaneously. Specifically, TMM transforms the original sentence and constructs a new multi-aspect sequence scheme, then apply multi-layer transformers to automatically learn to sentiments clues and potential relations of multiple aspects in a sentence. Compared with previous works that analyze one aspect each time, our TMM scheme not only helps improve computation efficiency but also achieves substantial improvements in the MAMS dataset. Finally, our method achieves the second place in NLPCC 2020 Shared Task 2 Evaluation. Experiment results and analysis also validate the effectiveness of the proposed method.

\subsubsection{Acknowledgements.}
This work was supported by the NSFC (No. 61976114, 61936012) and National Key R\&D Program of China (No. 2018YFB1005102).

%
%
%
\bibliographystyle{splncs04}
\bibliography{ref}

\end{document}